\documentclass[conference]{IEEEtran}
\IEEEoverridecommandlockouts

\usepackage{enumitem}
\usepackage{cite}
\usepackage{amsmath,amssymb,amsfonts}
\usepackage{algorithmic}
\usepackage{graphicx}
\usepackage{textcomp}
\usepackage{xcolor}
\usepackage{xfrac}
\usepackage{url}
\usepackage{listings}
\usepackage{booktabs}
\def\BibTeX{{\rm B\kern-.05em{\sc i\kern-.025em b}\kern-.08em
    T\kern-.1667em\lower.7ex\hbox{E}\kern-.125emX}}

\usepackage{tikz}
\def\layersep{2.5cm}

\newcommand{\R}{\mathbb{R}}
\newcommand{\argmax}{\operatorname{argmax}}
\newcommand{\fpu}{\mathsf{u}}

\begin{document}
	

\title{A Framework for Semi-Automatic Precision and Accuracy Analysis for Fast and Rigorous Deep 
Learning}

\author{\IEEEauthorblockN{Christoph Lauter}
\IEEEauthorblockA{\textit{Department of Computer Science - College of Engineering} \\
  \textit{University of Alaska Anchorage (UAA)}\\
  3211 Providence Dr
  Anchorage, AK, 99508\\
\url{christoph.lauter@christoph-lauter.org}\\
 ORCID: 0000-0001-7335-8220}
\and
\IEEEauthorblockN{Anastasia Volkova}
\IEEEauthorblockA{\textit{Université de Nantes}\\
	\textit{ CNRS, LS2N F-44000 Nantes, France}\\
\url{anastasia.volkova@univ-nantes.fr} \\
ORCID: 0000-0002-0702-5652}
}

\maketitle

\begin{abstract}
Deep Neural Networks (DNN) represent a performance-hungry application. Floating-Point (FP) and 
custom floating-point-like arithmetic satisfies this hunger. While there is need for speed, inference in 
DNNs does not seem to have any need for precision. Many papers experimentally observe that DNNs 
can successfully run at almost ridiculously low precision.

The aim of this paper is two-fold: first, to shed some theoretical light upon why a DNN's FP 
accuracy stays high for low FP precision. We observe that the loss of relative accuracy in the 
convolutional steps is recovered by the activation layers, which are extremely well-conditioned.
We give an interpretation for the link between precision and accuracy in DNNs.

Second, the paper presents a software framework for semi-automatic FP error analysis for the 
inference phase of deep-learning. Compatible with common Tensorflow/Keras models, it
leverages the frugally-deep Python/C++ library to transform a neural network into C++ code in order to 
analyze the network's need for precision. This rigorous analysis is based an Interval and Affine 
arithmetics to compute absolute and relative error bounds for a DNN. We demonstrate our tool with 
several examples.
\end{abstract}


\section{Introduction}

The area of Deep Learning (DL)~\cite{DL}, and in particular learning approaches based on Deep Neural 
Networks (DNNs), has seen some remarkable advances in the past decade. 
Neural networks are computing systems that can be seen as collections of connected nodes that are 
called neurons and are typically organized into sequentially inter-connected layers. 
Each layer performs an affine transformation defined by the layer's parameters (weights and bias), 
followed by a non-linear transformation called activation. 
DNNs can ``learn" to perform specific tasks by training on examples and then infer the results for new 
input data. For example, when 
given enough training samples, a classification network can learn the values of weights and biases for 
each neuron such that given a new image it can to distinguish cats from dogs on images. 

Many key algorithmic ideas underlying DNNs go back to as far as late 
1960s~\cite{minsky69perceptrons} 
with a reinstated interest in 1990s~\cite{tesauno}. 
The huge potential of DNNs is to solve complex problems while accepting input data in a raw and even 
heterogeneous form faced practical difficulties: hardware was not powerful enough, allowing only 
small-sized examples. 
It is not until the 2000s that some realistic problems could be 
solved by DNNs. The best example is the Imagenet~\cite{imagenet} image classification problem.
Then a huge wave of results in applying DL to real-world problems in the areas of computer vision, 
speech recognition, language-understanding and navigation in autonomous driving through 
reinforcement learning followed.  
Both increasingly large datasets and  increasingly complex models were critical for the success. 
For example, to recognize handwritten digits using MNIST network neural network is defined using 
around 0.7 
million parameters; the MobileNet network for the Imagenet problem requires around 27 million of 
parameters, while the BERT network goes up to 345 million parameters to solve Natural 
Language Processing problems. 

Deep Learning became an extremely computationally-hungry application in the end of the Moore's Law 
era~\cite{hennesy2019}, when again performance improvement in CPUs and even GPUs is not 
enough. 
%
On the computer arithmetic level, performance can be improved by reducing the bit-widths of data and 
operators, which results in smaller hardware area and memory-access times, and faster computations.
However, a tradeoff must be found in order to keep the DNN inference accurate.

While DNN training is usually performed in Floating-Point (FP) arithmetic using uniform float32 or 
mixed 
float32/half~\cite{nvidia_mp} precision, inference can be performed in smaller formats, or even in 
Fixed-Point arithmetic. 
There are several new low-precision FP formats that have been suggested by the major hardware 
manufacturers: 
bfloat16 (Intel~\cite{bfloat_Intel}, ARM~\cite{bfloat16_arm}), DLfloat (IBM)\cite{dlfloat}, MSFP8-11 
(Microsoft~\cite{msfp}) and the dedicated hardware is on its way.
Obviously, FPGAs and ASICs offer even richer design space.

Existing literature provides a large body of research on post-training quantization of 
DNN's parameters, typically down to 8-bit (Google's TPU) and 4-bit integers~\cite{4bitq}, or even down 
to 1 bit, which results in Binary Neural Networks~\cite{HubaraCSEB162, 
HubaraCSEB16}. 
Models can even be trained directly to have low-precision representation of weights and 
biases~\cite{Kravchik_2019_ICCV,choi2018, 
	CourbariauxB16c }. 

However, in the existing literature, the impact of rounding errors due to the precision of the underlying 
arithmetic has been, to the best of our knowledge, surprisingly missing. 
Perhaps negligible for close-to-float32 precisions, the arithmetic rounding errors in low-precision 
implementations can potentially grow and impact the network's classification choice.  
The experimental studies~\cite{bfloat_Intel, bfloat16_arm, bfloatStudy} make us think that DNNs 
can nevertheless maintain high inference accuracy even with low-precision FP arithmetic.
The first contribution of this paper is to shed some theoretical light on why this is the case by having a 
computer arithmetic look at DNN layers.

A typical study of the impact of rounding errors in DNNs is 
based on a comparison with a reference output on a (moderate) set of testing data. 
More formal approaches do exist, to analyze the robustness of DNNs with respect to 
perturbation of input parameters, e.g. the SafeAI 
project\footnote{http://safeai.ethz.ch}~\cite{AI2} based on abstract interpretation or 
SMT~\cite{ChangRG19}.
However, these tools do not account for FP rounding errors in their analyses. 
Our second contribution is to provide an interpretation of the impact of a precision choice upon the 
accuracy of a DNN.

Finally, we present a semi-automatic software framework for an automatic precision and accuracy 
analysis tool. Our versatile tool has a front-end accepting DNN models from common design 
frameworks such as Tensorflow/Keras, and is based on a generic error analysis technique, 
parametrizable by the target FP precision. The latter feature permits to analyze the behavior of DNNs 
upon a variety of target FP formats. 
In order to provide both relative and absolute error bounds, we introduce a combination of 
Affine and Interval arithmetics.


We start by recalling to the reader the basic notions of Deep Neural Networks in 
Section~\ref{sec:DNN}. Then, in Sections~\ref{sec:Arith} and~\ref{sec:DNNsAndArithmetic} we define 
our arithmetic toolkit and present a 
theoretical look at the numerical computations within DNN's layers, respectively. We follow by the 
description of the 
software tool and numerical experiments in Section~\ref{sec:TheTool} before concluding and 
discussing future research.

\section{Deep Neural Networks}\label{sec:DNN}


The basic computational units of DNNs are neurons, that can be seen as nodes parameterized by a 
weight 
$w\in \mathbb{R}$ and a bias $b \in \mathbb{R}$. Given an input $x \in \mathbb{R}$,  it computes an 
output $y = g(w\cdot x  + b)$, 
where $g: \mathbb{R} \rightarrow \mathbb{R}$ is a non-linear activation function. 
Typically, the inputs of a neural Network are assembled into a vector (e.g. a 32x32 pixel gray-scale 
image is a flattened into a 1024x1 vector), hence the per-layer computations are dot products. In a 
general case, DNNs operate on $N$-dimensional tensors. 

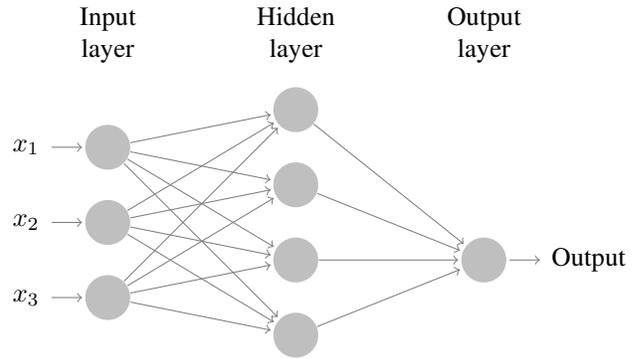
\begin{figure}

\begin{tikzpicture}[shorten >=1pt,->,draw=black!50, node distance=\layersep]
\tikzstyle{every pin edge}=[<-,shorten <=1pt]
\tikzstyle{neuron}=[circle,fill=black!25,minimum size=17pt,inner sep=0pt]
\tikzstyle{input neuron}=[neuron, fill=gray!50];
\tikzstyle{output neuron}=[neuron, fill=gray!50];
\tikzstyle{hidden neuron}=[neuron, fill=gray!50];
\tikzstyle{annot} = [text width=4em, text centered]

\foreach \name / \y in {1,...,3}
\node[input neuron, pin=left:$x_\y$] (I-\name) at (0,-\y) {};

\foreach \name / \y in {1,...,4}
\path[yshift=0.5cm]
node[hidden neuron] (H-\name) at (\layersep,-\y cm) {};

\node[output neuron,pin={[pin edge={->}]right:Output}, right of=H-3] (O) {};

\foreach \source in {1,...,3}
\foreach \dest in {1,...,4}
\path (I-\source) edge (H-\dest);

\foreach \source in {1,...,4}
\path (H-\source) edge (O);

\node[annot,above of=H-1, node distance=1cm] (hl) {Hidden layer};
\node[annot,left of=hl] {Input layer};
\node[annot,right of=hl] {Output layer};
\end{tikzpicture}
\caption{A DNN with 1 hidden layer. Each layer is parametrized by a weight matrix, a bias vector and an 
activation function.}\label{fig:DNN}
\end{figure}

Conceptually, DNN layers are divided into the input layer, the output layer and the intermediate, 
so-called hidden, layers as illustrated in Fig.~\ref{fig:DNN}. 
A trained DNN model is defined by its topology (number/type of layers) and the learned parameters 
(weights and biases).
Even though classically a network layer was comprised of a linear computation (e.g. dot product) and a 
non-linear activation function, modern literature often speaks of ``activation layer" as an independent 
entity. Following this trend, we assume that computational layers are  interleaved with the 
activation ones.

\paragraph*{Activation layers}
There exist a variety of non-linear activation functions, having different properties, e.g. 
bounded/unbounded, monotonic, continuously differentiable, etc.  
Some of the most common activations over vectors are the following ones: 
\begin{itemize}[wide]
	\item Sigmoid ($\operatorname{sigmoid}$): computes
	\begin{align}
	\sigma(x_i) = \frac{1}{1+e^{-{x_i}}}
	\end{align}
	\item Hyperbolic Tangent ($\operatorname{tanh}$): simply applies $\operatorname{tanh}(x_i)$ 

	\item Rectified Linear Unit ($\operatorname{ReLU}$): 
	\begin{align}
	\operatorname{ReLU}(x_i) = \max \{x_i, 0\}, \quad i=1,\ldots,n.
	\end{align}
	If the input $x$ is an $N$-dimensional tensor, all of the above functions are applied element-wise. 
	
	\item Softmax activation ($\operatorname{Softmax}$):  normalizes $x$ into a probability distribution 
	over output classes. It is evaluated via 
\begin{align}
 \operatorname{Softmax}(x_i)= \frac{e^{x_i}}{\sum_{j=1}^n e^{x_j}}, \quad i=1,\ldots,n.
\end{align}
If the input $x$ is an $N$-dimensional tensor, the $\operatorname{Softmax}$ activation is applied 
along each axis 
separately. 
\end{itemize}
It should be noted that all of the above functions are bounded, yielding output values in $[0,1]$, 
except for the $\operatorname{ReLU}$, which results in $[0; \bar{X}]$, i.e. it maintains an upper bound 
$\bar{X}$ on the layer's input while clipping negative values.

\paragraph*{Computational layers} Some of the most common layer types are:
\begin{itemize}[wide]
	\item Dense layer ($\operatorname{Dense}$): typically accepting an input data vector 
	$x 
	\in \mathbb{R}^{n}$ and parameterized by the weights matrix ${A} \in \mathbb{R}^{m \times 
	n}$ and a bias $b \in \mathbb{R}^{m}$. The output is compute via  ${y = A \cdot x + b \in 
	\mathbb{R}^m}$. 
	For multidimensional inputs, this operator extends into a tensor product. 
	
	\item Convolution\footnote{Should not be confused with a mathematical definition of convolution.} 
	layer ($\operatorname{Conv}$): its purpose is to extract the feature maps 
	out of data represented as multi-dimensional arrays through a linear transformation.
	This layer is parameterized by a convolution kernel that is 
	convolved with the layer input to produce a tensor of outputs. 
	The guide~\cite{dumoulin2016guide} offers a comprehensive description of the convolutional 
	arithmetic 
	for deep learning. For example, an image is given as a 3D tensor  $\operatorname{(rows, cols, 
	axes)}$, where $\operatorname{axes}$ provides the number of color channels. Convolution kernel is 
	also a 3D tensor of the size $\operatorname{(kernel\_width, kernel\_height, axes)}$.  
	Convolution operation consists of the kernel sliding across the input data 
	when at each location, the product between each element of the kernel and the input is 
	computed; consequently the results are summed up to obtain a scalar output in the current location. 
	The basic arithmetic operation in the convolution layers is again, the dot product. 

	\item Pooling layer ($\operatorname{Pool}$) : Pooling operations reduce the size of feature maps by 
	using some function to 
	summarize sub-regions, such as taking the average or the maximum value. 
	The basic 
	arithmetic operation in this layer is summation and/or $\max$ function. 
	
	\item Batch Normalization layer ($\operatorname{BatchNormalization}$): introduced 
	in~\cite{batchnorm}, 
	this technique is used to normalize the input of the layer and only then apply a dot-product. 
	 The idea is to divide the input data into mini-batches and first perform per-batch normalization. Let 
	 $m$ be a size 
	of a mini-batch $B$ and $x \in \mathbb{R}^{m \times d}$ be a $d$-dimensional input for the batch 
	normalization layer. Then, the normalization is applied for each dimension separately via:
	\begin{align}
	\tilde{x}^{(k)}_i = \frac{x_i^{(k)} - \mu_B^{(k)}}{\sqrt{\sigma_B^{(k)^2} + \epsilon}},  \quad k=1,\ldots, d, 
	\; 
	i=1,\ldots,m,
	\end{align} 
	where the mean $\mu_B = \frac{1}{m}\sum_{i=1}^{m} x_i \in \mathbb{R}^k$ and variance 
	$\sigma_B^{2} 
	=  
	\frac{1}{m}\sum_{i=1}^{m} (x_i - \mu_B)^2 \in \mathbb{R}^k$ are computed per mini-batch 
	$B$ and $\epsilon > 0$ is a small  parameter. After normalization, the input vector is transformed as 
	follows : $y_i^{(k} = \gamma^{(k)} 
	\tilde{x}_i^{(k)} + \beta^{(k)}$, where $\gamma$ and $\beta$ are $d$-dimensional vectors of  
	parameters that are learned during the optimization. 
\end{itemize}
The continuously updating  list of computational layers can be 
found in the documentation for the Tensorflow/Keras.

\section{Arithmetic toolkit}\label{sec:Arith}
FP operations, such as addition, subtraction,
multiplication or more complex functions like $\exp, \tanh$, can
necessarily not be exact: floating-point representation uses finite
memory. Rounding is hence necessary after almost every
operation. These roundings induce error into the computation,
affecting the final result~\cite{Handbook}. For a DNN, this means
e.g. that the output class and the attached confidence probability are
affected by rounding error. When confidence is low, that error might
have even toggled the class the DNN output. In order to make DNNs
rigorous, the overall FP rounding error affecting the
results must be analyzed.

The error due to solely one FP operation mainly depends on the
\emph{precision} $k$ of the FP format used. For
binary FP formats --which we shall focus on in this paper-- precision
expresses the number $k$ of bits held in the format's mantissa. For
example, for the IEEE754-2019 binary32, $k=24$ and for IEEE754-2019
binary64, $k=53$. For an IEEE754-2019 FP operation with precision $k$
that does not overflow nor underflow, i.e. exceed the format's
exponent range, the following holds for
rounding-to-nearest~\cite{Handbook}: let $\fpu = 2^{-k+1}$. Then, for
every FP input $a,b$, there exists $\varepsilon \in
\left[-\sfrac{1}{2};\sfrac{1}{2}\right]$ such that
\begin{equation}\label{eq:firstFPmodel}
  a \odot b = \left( a \circ b \right) \cdot \left( 1 + \varepsilon\,\fpu \right)
\end{equation}
where $\odot$ is the FP realization of operation $\circ \in \left
\lbrace +, -, \times, / \right \rbrace$. A similar bound is available
for unary operations such as $\sqrt{~}, \exp, \tanh$. This
representation of the FP rounding error is also called the first FP
error model~\cite{High02}. Remark that~\eqref{eq:firstFPmodel} holds
independently of the value of $\fpu$, i.e. independently of the chosen
precision $k$. This model therefore allows code to be analyzed for
a given precision so to tailor it for an application.

The difficulty in analyzing a given FP code's \emph{accuracy} by
analyzing the total amount of error affecting a result value lies in
the intricate ways the different elementary errors, which are given by
eq.~\eqref{eq:firstFPmodel}. For example, suppose two multiplication
operations are execute one after the other. By applying
eq.~\eqref{eq:firstFPmodel} twice, we already obtain:
\[
a \otimes \left( b \otimes c \right) = \left(a \times b \times
c\right) \cdot \left( 1 + \varepsilon_1 \, \fpu + \varepsilon_2 \,
\fpu + \varepsilon_1 \, \varepsilon_2 \, \fpu^2 \right).
\]
A single scalar product of a DNN's convolution layer with $n$ inputs
uses $n$ multiplications and $n$ additions, each of which will result
in an error term $\varepsilon_i \, \fpu$ by application of
eq.~\eqref{eq:firstFPmodel} but will also combine with all other error
terms in the most intricate way. Manual analysis with this approach is
hence completely intractable. Wilkinson and Higham therefore invented
extended ways of analyzing the combination of elementary errors in FP
code~\cite{Wilkinson, High02}. However, their analysis requires human
insight into the different algorithms used, such as addition of FP
values, scalar products etc. When existing code is to be analyzed for
accuracy automatically using software, Higham's approach is hence not
usable either.

A workable approach is found with \emph{Affine Arithmetic} (AA), as
developed e.g. by Putot~\cite{putot2011}. Every FP quantity
$\hat{q}$ is annotated with a bound $\overline{\varepsilon}$ that
expresses the quantity's error with respect to the mathematically
ideal, but unknown quantity $q$ in a similar manner as in
eq.~\eqref{eq:firstFPmodel} above:
\begin{equation}\label{eq:affineRelative}
  \hat{q} = q \cdot \left( 1 + \varepsilon \, \fpu \right) \quad
  \text{with} \left \vert \varepsilon \right \vert \leq
  \overline{\varepsilon}.
\end{equation}
If $\hat{q}$ stems from an exact quantity $q$ in a way that involves
only one FP operation, $\overline{\varepsilon}$ is set to
$\sfrac{1}{2}$ according to~\eqref{eq:firstFPmodel}. Otherwise,
when $\hat{q}$ is the result of combining two quantities $\hat{r}$ and
$\hat{s}$ with an operation $\odot$, the error terms
$\overline{\varepsilon}_r$ and $\overline{\varepsilon}_s$ as well as
the error term $\overline{\varepsilon}_\odot$ due to the rounding in
the operation $\odot$ are combined to yield one single new error term
$\overline{\varepsilon}$ for $\hat{q}$ with respect to $q$. This
combination is specific to each operation type. For example, for
addition, $\odot = \oplus$ we obtain:
\begin{eqnarray}\label{eq:affineAddition}
  \hat{q} & = & \hat{r} \oplus \hat{s}  = \left( \hat{r} + \hat{s} \right) \cdot \left( 1 + \varepsilon_\odot \, 
  \fpu \right)  \nonumber \\
  & = & \left( r \cdot \left( 1 + \varepsilon_r \, \fpu \right) + s \cdot \left( 1 + \varepsilon_s \, \fpu \right) \right) \cdot \left( 1 + \varepsilon_\odot \, \fpu \right) \nonumber \\
  & = & \left( r + s \right) \cdot \left(1 + \varepsilon_r \, \frac{r}{r + s} \, \fpu + \varepsilon_s \, \frac{s}{r + s} \, \fpu \right) \cdot \left( 1 + \varepsilon_\odot \, \fpu \right) \nonumber \\
  & = & q \cdot \left( 1 + \varepsilon \, \fpu \right),
\end{eqnarray}
with
\begin{equation} \label{eq:affineAdditionErrorCombine}
  \varepsilon = \frac{\left(1 + \varepsilon_r \, \frac{r}{r + s} \, \fpu + \varepsilon_s \, \frac{s}{r + s} \, \fpu \right) \cdot \left( 1 + \varepsilon_\odot \fpu \right) - 1}{\fpu}.
\end{equation}
To annotate $\hat{q}$ with $\overline{\varepsilon}$ bounding
that $\varepsilon$ in~\eqref{eq:affineAdditionErrorCombine}, we
need to
\begin{itemize}
\item have access to the annotations $\varepsilon_r$ and
  $\varepsilon_s$ of the operands,
\item know a bound on $\varepsilon_\odot$, which is provided by
 ~\eqref{eq:firstFPmodel} and
\item to be able to bound the error amplification (or attenuation)
  quantities $\alpha_r = \frac{r}{r+s}$ and $\alpha_s = \frac{s}{r+s}$.
\end{itemize}
For the latter task, to bound $\alpha_r$ and $\alpha_s$, we use
Interval Arithmetic (IA), as we shall explain below. We have detailed only 
the case of FP addition, $\odot = \oplus$. Similar error combination
rules exist for the other operations that occur in DNNs, such as
subtraction, multiplication, division, square root, $\exp$, $\tanh$.

However, the AA approach above is based on a relative error model: the
term $\varepsilon \, \fpu$ describes the relative error of the FP
quantity $\hat{q}$ with respect to the ideal, unknown quantity
$q$. Such a relative error bound does not always exist, in which case
relative AA breaks down. An easy-to-understand example is when the
result of an FP addition $\hat{r} \oplus \hat{s}$ cancels out
completely, amplifying the incoming errors $\varepsilon_r \, \fpu$ and
$\varepsilon_s \, \fpu$ by an infinite amount. Typically, in this
case, the quantities $\alpha_r$ and $\alpha_s$ do not stay bounded as
their denominator becomes zero.

A solution to this issue is to use AA not with a relative error term
but with an absolute error term. An FP quantity $\hat{q}$ is labeled
with an absolute error bound $\overline{\delta}$ such that
\begin{equation}\label{eq:affineAbsolute}
  \hat{q} = q + \delta \, \fpu \quad
  \text{with} \left \vert \delta \right \vert \leq
  \overline{\delta}.
\end{equation}
For FP addition, the absolute error terms just add up, plus an
additional absolute error term, which can easily be deduced out of the
relative term given by~\eqref{eq:firstFPmodel}, by multiplying the
relative error bound by an upper bound on the exact result's absolute
value. Such an upper bound is easy computed using IA

As a matter of course, due to the ``there's no free lunch''-rule, the
ease of use of absolute AA for addition comes with issues for other
operators like division, where amplification terms
similar to $\alpha_r$ and $\alpha_s$ become unbounded and hurt
absolute AA the same way as addition hurts relative AA.

Our solution to this issue is to maintain both absolute and relative
error ``bounds'' $\overline{\delta}$ and $\overline{\varepsilon}$ for
each quantity $\hat{q}$ and to let them become infinite whenever no
such bound exists. Operators like addition, multiplication, division,
square root, $\exp$ etc. try to propagate both the absolute and the
relative error bounds whenever possible, using the information in both
bounds when appropriate. This is, addition and subtraction, which may
cancel, propagate the absolute error bound and may yield an infinite
relative error bound. Multiplication, division and square root start
off the relative error bounds and propagate those. Exponential propagates
the entering absolute error bound as a relative error bound as in
\[
e^{q + \delta\,\fpu} = e^q \cdot \left( 1 + \frac{e^{\delta \, \fpu} - 1}{\fpu} \, \fpu \right).
\]
Logarithm does the inverse, transforming a relative error bound into
an absolute one. The function $\tanh$, used a lot in DNNs activation
layers, can propagate the absolute error with no amplification factor
and may propagate the relative error bound $\overline{\varepsilon}$
with a small amplification factor of $2.63$ whenever
$\overline{\varepsilon} \, \fpu \leq \sfrac{1}{4}$. The details of
this combined absolute and relative AA (CAA) go beyond
the scope of this paper but we may state:
\begin{eqnarray*}
  \tanh\left(q + \delta \, \fpu\right) & = & \tanh(q) + \delta' \, \fpu \quad \text{with } \overline{\delta}' = 
  \overline{\delta}, \\
  \tanh\left(q \cdot \left( 1 + \varepsilon \, \fpu \right) \right) & = & \tanh(q) \cdot \left( 1 + \varepsilon' \, \fpu \right) \\ & & \quad \text{with } \overline{\varepsilon}' = 2.63 \, \overline{\varepsilon} \text{ if } \overline{\varepsilon} \, \fpu \leq \sfrac{1}{4}.
\end{eqnarray*}
Whenever possible, the proposed CAA improves the one bound --absolute
or relative-- of a quantity using the other. For example, it is often
possible to deduce tight relative error bounds out of absolute when a
quantity can be shown never to be zero. Likewise, an absolute error
bound is readily deduced from the relative one and an upper bound on
the quantity.

In order to do so and, as explained above, to be able to combine the
operands' error terms and to bound absolute elementary errors using
eq.~\eqref{eq:firstFPmodel}, bounds for the quantities occuring in a
computation must be known. We compute these bounds using \emph{Interval
Arithmetic} (IA)~\cite{Moore}. For IA, each quantity is replaced by an
interval the quantity can be shown to lie in. Each IA operator working
on intervals produces an interval that surely encompasses all possible
images of the operation and operands in the operand intervals. All
roundings are performed in such a way --viz. outwards-- that this enclosure
property is satisfied even in the presence of roundings~\cite{Moore}.

Both IA and (absolute, relative or combined) AA are plagued by a
phenomenon called the \emph{decorrelation effect}~\cite{putot2011}. Consider the following code 
snippet:
\begin{lstlisting}[language=C]
  y = x;
  z = x - y;
\end{lstlisting}
While mathematically, $z$ will always be zero, as $y=x$, and while
even IEEE754-2019 FP code ensures that \lstinline[language=C]{z} will
be zero due to full cancellation of all bits of
\lstinline[language=C]{x} and its copy \lstinline[language=C]{y}, IA
and AA will have no global understanding that $x$ and $y$ are
correlated, and --actually-- equal. So assuming $x$ to be bounded by
the interval $\left[ -1;1 \right]$, $z$ will evaluate to the interval
$\left[-2;2\right]$ instead of the interval $\left[0;0\right]$. For
CAA, the issue will be that the relative error bound
$\overline{\varepsilon}$ on $\hat{z}$ instead of becoming zero as the
errors on $\hat{x}$ and $\hat{y}$ cancel out will become infinite due
to the detected ``catastrophic'' cancellation. The absolute error
bound $\overline{\delta}$ for $\hat{z}$ will not become zero but the
double of the one on $\hat{x}$. For all kind of arithmetic techniques,
such as IA or CAA, which only label quantities in code with interval
bounds or absolute and relative error bounds but which do not gain any
global understanding of the code, the decorrelation effect has no
simple solution. It depends on the application whether or not the
decorrelation effect occurs and whether or not its consequences are
bearable or not for that type of application. As we shall see in more
detail in Section~\ref{sec:DNNsAndArithmetic}, in code for DNNs, the
decorrelation effect does occur, typically in precisely the way
illustrated with the code sequence above. It does not occur in its
even more intractable appearances, like in cases when $y = \sin x$ and
$z = x - y$, where $y$ and $x$ are correlated for small $x$, as the
Taylor series of $\sin$ starts with $x - \sfrac{1}{6} \, x^3$.

For the easy case when two variables in code correlate because they
are copies on of the other --as in the example code illustrated
above-- a simple solution to overcome the decorrelation effect in CAA
(and IA) exists: all FP quantities analyzed by CAA are labeled with a
unique identifier that relates to their moment of creation in the
execution of the program. This identifier is never repeated for any
other FP quantity but for assignment, where the identifier does get
copied. Subtraction (and division) operations in CAA can then start by
checking whether the identifier of both operands happens to be the
same. If it does, both operands are correlated as they are copies one
of each other. Interval bounds of $\left[0;0\right]$ and CAA error
bounds of $\overline{\delta} = 0$ and $\overline{\varepsilon} = 0$ can
then be returned. This solution is crude but addresses all simple
decorrelation cases found in DNNs code.

Yet another issue with code analysis with CAA and IA comes in the form
of \lstinline[language=C]{if} statements depending on FP variables
--which are to be analyzed-- and, generally, \emph{control flow} depending on FP
variables~\cite{putot2011}. FP code might take the one or the other branch
by evaluating a comparison like \lstinline[language=C]{x < y} to a
boolean, which, of course, might be falsed due to the errors on $x$
and $y$. In contrast, CAA and IA, which replace $x$ and $y$ by whole
classes (intervals resp. abstract approximate quantities with bounded
error measured in units of $\fpu$), cannot even evaluate the
expression to a unique boolean in cases where the intervals for $x$
and $y$ intersect or where the errors make the boolean answer not
unique. Some approaches for this control flow issue have been
proposed~\cite{fluctuat}.

Fortunately, in code for DNNs, this issue is virtually absent. Code
for DNNs, in inference mode, does not contain control flow in the form
of loops that depend on FP values. In other words, no iterative FP
techniques are used. All control flow for loops comes from the DNN's
configuration and the respective dimensions of the manipulated
tensors. As we shall see in more detail in
Section~\ref{sec:DNNsAndArithmetic}, the only
\lstinline[language=C]{if} statements in DNNs that depend on FP values
are encountered in activation layers such as pooling or softmax
layers. This is due to the very nature of DNNs: in order to make
training possible, from a bird-view perspective, DNNs need to
represent (non-linear) functions that are differentiable. Branches would necessarily introduce 
discontinuities of
that derivative. In the concerned code sequences, the
\lstinline[language=C]{if} serve the only purpose of computing minima
and maxima on vectors of FP values, hence does not influence the output directly.
We solve the control flow issue in a similar way as the decorrelation effect: the point is to
provide the CAA and IA arithmetics, which, again, are concerned with
local effects, with just enough global insight on the program's
logic. For instance, the quantities analyzed with CAA and IA can be
labeled with bounds given in the form of other CAA+IA quantities that
are minimum or maximum bounds for them. Subsequent CAA or IA
operations, like subtraction, can then exploit the fact that if for example a quantity
$\hat{q}$ is upper-bounded by $\hat{M}$, i.e. $\hat{q} \leq \hat{M}$,
the result of the subtraction $\hat{q} - \hat{M}$ will always be
bounded by $\hat{q} - \hat{M} \leq 0$. 

As a matter of course, DNNs that perform classification tasks do
contain \lstinline[language=C]{if} statements that depend on FP
values. These statements are the one executed as the very last step,
when the one-hot output of a softmax
layer~\cite{DL} gets translated into the predicted
numerical integer class. This code boils down to computing the integer
$\argmax$ index for a vector of FP values of probabilities; the DNN
picks the class that is the most probable. However, it is the very aim
of this paper to analyze the effects of roundings in the FP arithmetic
for the DNN's inference on the output class, which we discuss in
Section~\ref{sec:DNNsAndArithmetic}.

To wrap it up, our approach is to analyze DNNs for FP
rounding errors using CAA and IA, where the FP quantity in the DNN to be
analyzed gets replaced by an arithmetical object containing the following entries:
\begin{itemize}[wide]
\item a unique ID of the quantity, in the form of an integer,
\item the FP value in the IEEE754-2019 (or any other) FP format that
  would be used if the DNNs were implemented without this enhanced
  CAA+IA arithmetic,
\item an interval holding the actual error of the latter FP value, for
  reference purposes,
\item an absolute error bound $\overline{\delta} \in \R^+ \cup \left
  \lbrace +\infty \right \rbrace $, for this quantity, in units of $\fpu$,
\item a relative error bound $\overline{\varepsilon} \in \R^+ \cup
  \left \lbrace +\infty \right \rbrace $, for this quantity, in units of
  $\fpu$,
\item an interval safely enclosing all possible values for this
  quantity if no FP rounding error occurred,
\item an interval safely enclosing all possible values for this
  quantity, as it is evaluated with rounding FP arithmetic and,
\item optionally, a lower and an upper bound for this quantity. These
  bounds are given in the form of arithmetical objects of the same
  nature.
\end{itemize}
All operators required for DNNs, starting with assignment, going over
computational operators like $+, -, \times, /, \sqrt{~}$ to functions
like $\exp$, $\log$ and $\tanh$ are overloaded to work on such CAA+IA
arithmetical objects, propagating all entries as described above. As a
result, a DNNs run on an example input, widened with interval bounds
for the inputs' ranges, provides an output in these arithmetical
objects, from which errors on probabilities etc. can be read off. As
the absolute and relative error bounds are expressed in units of
$\fpu$, that same output can be used to tailor a DNN's FP precision to
just the right amount of tolerable final error. We shall describe the
use of this arithmetic just below, in
Section~\ref{sec:DNNsAndArithmetic}.  As for the technical realization
of this enhanced CAA+IA arithmetic in an actual software tool, we
refer the reader to Section~\ref{sec:TheTool}.

\section{Computer arithmetic look at DNNs} \label{sec:DNNsAndArithmetic}
As we shall see in the next Section~\ref{sec:TheTool}, the enhanced
CAA arithmetic we just described is able to automatically analyze
given FP code for DNNs and to come up with absolute or relative error
bounds, expressed in units of $\fpu$, that are pretty tight and suit
their purpose. However, as useful as this automatic analysis might be
for application programmers, we wanted to ensure its tightness and
validity from a more theoretical standpoint. This Section strives at
providing this insight. For the sake of brevity, we shall focus on
DNNs for classification problems. The analysis is similar for other
types of problems. We will present a concrete example of a DNN for a
non-classification problem in Section~\ref{sec:TheTool}.

DNNs for classification problems transform high-dimensional input data
into an output vector that is a one-hot representation of the class
detected for the input's class. This is, the output vector has as many
entries as there are classes, and the $i$-th entry of that vector
contains a probabilistic estimate of the confidence of the DNN the
input is in the $i$-th class. That estimate is expressed as a
probability; all entries are hence between $0$ and $1$ and sum up to
$1$. Post-processing after a DNN picks the class for which the
confidence estimate is highest, computing the $\argmax$ on the output
vector. The index of this output class can then be translated into
e.g. a textual representation of the class, such as \emph{``Cat''} or
\emph{``Dog''}.

In the case when the maximum confidence estimate is at $50\%$ and the
second-to-maximum confidence estimate is also at $50\%$, the slightest
change to these output values will of course make the DNN commit a
misclassification, outputting e.g. \emph{``Dog''} when the input
represents a \emph{``Cat''}. Such a slight change may stem from FP
rounding errors. For DNN input data where maximum confidence is at
$50\%$, no FP arithmetic --but exact arithmetic-- exists avoiding
misclassifications. However, when external knowledge on the DNN exists
that guarantees that, on all possible inputs\footnote{For a reasonable
  definition of what a \emph{possible} input is.}, the DNN will output
a one-hot vector with a top-1 value $p^\star > 0.5$, it can be guaranteed
that the second-to-maximum, top-2, value will be $p^\dagger < 1 - p^\star$,
leaving a margin of $\sfrac{1}{2} \, \left( p^\star - p^\dagger
\right) > p^\star - \sfrac{1}{2}$ for each of the maximum and
second-to-maximum entry to be affected by FP rounding error. Gaining
this external knowledge is beyond the scope of this article, but
approaches like SafeAI~\cite{AI2} seem to be able to provide it. 
This external minimum bound may also just be specified,
accepting a certain percentage of misclassifications.

From a computer arithmetic perspective, we may hence assume that there
is an absolute FP error margin $\mu = \underline{p}^\star -
\sfrac{1}{2}$ available for each element in the output vector, where
$\underline{p}^\star > 0.5$ is the minimum bound established with
external knowledge. Similarly, we may assume a relative FP error
margin $\nu = \frac{2\,\underline{p}^\star - 1}{2\,\underline{p}^\star
  + 1}$ available. Our job is to rigorously ensure that no
misclassification may occur given that error margin. Hence we need to
choose FP precision $k$, resp. $\fpu = 2^{-k+1}$ in such a way that we
can guarantee that the DNN's inference accuracy is enough so
that the FP rounding error does not exceed the margins. We may hence
start reclimbing the DNN's FP algorithm from its end with that margin
as some kind of FP error budget to be burnt for FP roundings.

As their last layer, most classification DNNs have a $\operatorname{Softmax}$ layer, as
it was defined in Section~\ref{sec:DNN}. We must hence analyze the FP
error in output of a $\operatorname{Softmax}$ layer. This analysis will also serve as
an illustration of error analysis for the different layers; we
analyzed all layers, for the sake of brevity, we shall only report on
the $\operatorname{Softmax}$ layer. The error in output of a $\operatorname{Softmax}$ layer has two
sources: (1) the FP rounding errors committed during the layer's
evaluation and (2) the errors present in input to the layer,
propagated in an amplified or attenuated manner by the layer. The
analysis of the first kind of errors, the rounding errors, is trivial,
as the $\operatorname{Softmax}$ function required just the evaluation of exponentials,
a division (often implemented as a division of a logarithm) and the
summation of positive values, obtained by exponentiating the input. We
shall hence not address this point any further.

For the propagated error of a $\operatorname{Softmax}$ layer, the following analysis
can be performed; herein, $\hat{x}_i$ are the elements of the computed input vector affected by an
absolute error $\delta_i$, approximating the unknown, ideal $x_i$. The $\hat{y}_i$ are the
output vector elements, approximating the unknown, ideal $y_i$.
\begin{eqnarray}
  \hat{y}_i & = & \frac{e^{x_i + \delta_i}}{\sum\limits_k e^{x_k + \delta_i}} \nonumber \\
  & = & \frac{e^{x_i}}{\sum\limits_k e^{x_k} \, \cdot \, \left( 1 + \frac{\sum\limits_k e^{x_k} \cdot \left( 
  e^{\delta_k - \delta_i} - 1 \right)}{\sum\limits_k e^{x_k}} \right)} \nonumber \\
  & = & y_i \cdot \left( 1 +   \frac{1}{1 + \eta_i} - 1 \right) = y_i \cdot \left( 1 + \varepsilon_i \right) 
\end{eqnarray}
with
\[
\eta_i = \sum\limits_k \frac{e^{x_k}}{\sum\limits_j e^{x_j}} \cdot \left( e^{\delta_k - \delta_i} - 1 \right).
\]
This quantity is easily bounded with
\begin{eqnarray*}
  \left \vert \eta_i \right \vert   & \leq & \sum\limits_k \frac{e^{x_k}}{\sum\limits_j e^{x_j}} \cdot 
  \max\limits_t \left \vert e^{\delta_t - \delta_i} - 1 \right \vert \\
  & \leq & \max\limits_k \left \vert e^{\delta_k - \delta_i} - 1 \right \vert.
\end{eqnarray*}
With some mild assumptions on bounds for the $\delta_k$ and $\eta_i$, it
can further be shown that the relative error $\varepsilon_i$ affecting $\hat{y}_i$
is bounded by
\begin{equation}\label{eq:softmax}
  \left \vert \varepsilon_i \right \vert \leq \sfrac{11}{2} \, \max\limits_k \left \vert \delta_k \right \vert,
\end{equation}
essentially by taking the Taylor development of $e^x - 1$.

This analysis therefore shows us the following: the $\operatorname{Softmax}$ layer
transforms the absolute error in its input into a relative error of
approximately\footnote{i.e. $5.5$ times larger} the same amount in output. Our margin of $\nu
= \frac{2\,\underline{p}^\star - 1}{2\,\underline{p}^\star + 1}$
becomes hence an absolute error margin in input of the $\operatorname{Softmax}$ layer.
This input of the $\operatorname{Softmax}$ layer is in general the output of a
convolutional layer, where the arithmetical
difficulty is in a summation, which lives very well when it just needs
to satisfy an absolute error bound. Amazingly, the bound given
above does not at all depend on the number of elements of
the vectors $x$ and $y$.

In order to illustrate this stability argument more intuitively, let us give a
numerical example: let $\underline{p}^\star = 0.60$, i.e. the
classifying DNN shows at least $60\%$ confidence for the best output
class. Then $\nu > 0.0909 > 2^{-3.45}$, meaning that FP results with
about $3.45$ valid bits are sufficent. Then a
maximum element-wise absolute error of $\frac{0.0909}{5.5}>1.65\cdot10^{-2}$ is still tolerated on the 
input of the
softlayer. This means for a convolution or dot-product,
i.e. summation, fixed-point arithmetic with a quantization unit of
$1.65\cdot10^{-2}$, i.e. about $2^{-6}$ is enough. FP arithmetic can
only do better, its precision is at least these $6+g$ bits, provided
the inputs to the summation are bounded around $2^g$.

The boundedness of the values manipulated by DNNs is something we have
already stated in Section~\ref{sec:DNN}. Most activation layers bound
their outputs to $\left[ 0; 1 \right]$, equivalent to a bound $2^g =
2^0$. Convolutional or fully-connected layers also exhibit pretty
small bounds $2^g$, which are easily established, and, by the way,
perfectly bounded with IA~\cite{Shary}, considering the boundedness of
the DNN's coefficients, the relatively small dimensions of the
manipulated vectors, matrices and tensors and the boundedness of the
preceeding input.

It is hence all but surprising to observe that 
\begin{itemize}[wide]
\item DNN inference behaves very well for FP arithmetic with low precision,
\item DNN inference behaves very well for FP arithmetic with low exponent range,
  as fixed-point arithmetic already provided enough accuracy for the subsequent layers, such as softmax,
\item and analysis with CAA that exhibits small FP error bound in output does provide tight and sensible results.
\end{itemize}

\begin{table*}[]
	\centering 
	\begin{tabular}{@{}lccccclcc@{}}
		\toprule
		&  & max absolute error in $\fpu$ && max relative error in $\fpu$ &  & analysis 
		time &  & required 
		precision to prevent misclassificaton with $\underline{p}^\star = 0.60$ \\ \cmidrule(lr){3-3}  \cmidrule(lr){5-5} \cmidrule(lr){7-7} 
		\cmidrule(l){9-9} 
		Digits    &   & $1.1\fpu$ &&  $3.4\fpu$                  &  &            12s per  class               &  & $k=8$
		                                          \\
		MobileNet &  &           $22.4\fpu$         & &        $11.5\fpu$              &  &            4.2h per  
		class                    &  
		&                    $k=8$                      
		\\
		Pendulum  &  & $1.7\fpu$                   &  &         --         &  &            100ms                & 
		&                                           -- 
		\\ \bottomrule
	\end{tabular}
	\vspace{\baselineskip}
	\caption{Numerical results for experiments with $\fpu \leq 2^{-7}$. 
	}
	\label{tab:results}
\end{table*}

\section{CAA-based FP Error Analysis and Experimental Results} \label{sec:TheTool}

We implemented our semi-automatic FP accuracy analysis tool building
upon a combination of existing software packages for DNNs, such as
frugally-deep\footnote{\url{https://github.com/Dobiasd/frugally-deep}},
which we patched pretty heavily. We coupled these packages with a C++
implementation of the enhanced CAA that we have described in
Section~\ref{sec:Arith}. This C++ implementation of CAA was written
from scratch. The implementation is currently based on IA provided by
MPFI 1.5.3, which is itself based on MPFR 4.0.2 on top of GMP
6.1.2~\cite{MPFI, MPFR, GMP}. However, we wrapped MPFI in a C++ façade
class in order to facilitate transition to other IA libraries later.
 We use {\tt g++} version 8.3.0 to
compile our code. The frugally-deep library we are using requires the
use of C++ in its C++20 version~\cite{CPP20}. Our contribution in
terms of code consists in C++ classes to implement CAA as we have
described it and in the patches required to allow for binding and use
of that CAA arithmetic instead of plain IEEE754-2019 arithmetic in
frugally-deep. Our workflow runs only with our version of
frugally-deep, not with a stock version. Thanks to frugally-deep, our
semi-automatic FP accuracy analysis tool is compatible with almost all
 DNNs as they are designed and trained with
Tensorflow/Keras~\cite{Tensorflow, Keras}.
The frugally-deep package first converts DNN models to JSON files, and then provides C++ header 
classes that allow loading of JSON files as object graphs that can be evaluated on the input data.
  The
  frugally-deep library leverages several other C++ libraries for
  this task, the most prominent of which is Eigen~\cite{Eigen} that permits binding of custom 
  arithmetic.
%


Our CAA class structure consists of three classes: a façade class for
the front-end binding with frugally-deep; a class that actually implements the CAA arithmetic and 
overloads all necessary arithmetic operations; and a back-end wrapper for IA. 
We did not use existing MPFI wrappers in C++,
in order to have the possibility to exploit the performance advantage
of new C++20 features, like move constructors and move assignment
operators.

Our workflow runs as follows: using frugally-deep we construct a C++ program to load a DNN model 
designed in Tensorflow/Keras, as well as the input data, expressed with CAA objects. The bounds on 
these data are trivial in most cases, e.g. image data gets annotated with 8-bit unsigned values in $\left[ 
0; 255 \right]$.
We run the
resulting program for all possible classes to cover all possible
control flows. And this can be done only for one representative of the class, no additional tests are 
required. The program outputs the inference result and the absolute and relative
error on it. The error bounds are all given in units of $\fpu$, an upper bound on which is 
user-configurable. The output error bounds can
then be used to tailor the DNNs actual FP arithmetic, by applying the
theory we described in Section~\ref{sec:DNNsAndArithmetic}, determining the value of $\fpu$
such that the required accuracy bounds are still met.

We demonstrate our tool on several examples of DNNs and give some results in 
Table~\ref{tab:results}.
\paragraph{Digits} We built a simple DNN for the recognition of hand-written digits and trained it on 
the MNIST dataset~\cite{lMNIST}. This model requires around 0.7 million parameters and consists of 
three 
$\operatorname{Dense}$, two  $\operatorname{ReLU}$ and a $\operatorname{Softmax}$ 
layer. As input, it takes $28\times28$ gray-scale images (i.e. a flattened vector of length $784$) and 
has a $10$-dimensional output vector whose $i^{\text{th}}$ element indicates the probability that the 
input 
image is the digit $i$. Table~\ref{tab:results} illustrates the results of analysis, where the maximum 
absolute and relative errors denote the maximum errors over all possible classes. 
We also observed that on the top-1 choice, the relative error bounds are quite tight, while on the other 
elements the relative error looks less good. However, the bound~\eqref{eq:softmax} still holds and in 
any case, the absolute error stays low. Our analysis shows that the network can safely run with 7-bit 
precision FP.
\paragraph{MobileNet} We used a Keras pre-trained model for this considerably bigger network for 
Imagenet classification. MobileNet requires around 27 million parameters. The complete architecture 
can be found in Keras documentation, we will only state here that it is a Convolutional Neural Network 
with 28 $\operatorname{Conv}$ layers, 27 $\operatorname{BatchNorm}$ layers followed by 
$\operatorname{ReLU}$ activations and a $\operatorname{Softmax}$ layer that classifies $224\times 
224$ 
RGB images over 1000 classes.
This challenging example revealed a performance bottleneck in our tool.
To analyze the model over one class it took the tool around 4 hours on a conventional laptop.
Our performance analysis determined that most of the analysis time was dedicated to the memory 
allocation process somewhere deep in MPFI. 
Regardless of the analysis time, the tool successfully illustrated that even for large-scale models our 
analysis techniques compute tight error bounds.
\paragraph{Pendulum} This small neural network model comes from the context of reinforcement 
learning applied for autonomous  control~\cite{ChangRG19} and aims at approximating a Lyapunov 
function for a non-linear controller. The article~\cite{ChangRG19} proposes a new methodology for 
certified approximation of neural controllers based on SAT theory. This example is interesting 
from the formal verification point of view: our bound on the absolute error can be effortlessly 
incorporated into the existing verification procedure. This network has two $\operatorname{Dense}$ 
layers and two $\operatorname{tanh}$ activations. It takes a $2D$ coordinate vector as 
input and, as in~\cite{ChangRG19}, we tested it for the interval $[-6; 6]$. Our tool provided an 
absolute error bound  in a fraction of a second. A relative error bound does not exist since the output 
interval contains zero.



\section{Conclusion and Perspectives}\label{sec:conclusion}

With this work we proposed a semi-automated way to \textit{bound} and \textit{interpret} the impact of 
rounding errors due to the precision choice for inference in generic DNNs. 
We presented a software tool that, thanks to frugally-deep library, can receive 
any TensorFlow/Keras model in the front-end. We support the most common activation and 
computational layers. 

The back-end that we developed automatically computes and propagates rounding errors through the 
computations. 
For this, we have introduced a combination of Affine and Interval Arithmetics called CAA. 
This new construction permitted us to compute both relative and absolute error bounds.
Our implementation of CAA is based on rigorous error analysis for arithmetic operations, as 
well as for all the necessary elementary functions for activation layers (e.g. $\operatorname{tanh}$, 
$\operatorname{exp}$). 
We enhance CAA with just enough global insight on the program's logic in order to fight the 
decorellation 
effect and take care of the control flow depending on FP variables. 
Software implementation of the arithmetic is done in C++ in a generic way.

We offer a computer arithmetic look at the computations with the DNNs.
When analyzing the computations within activation layers, we establish that activation functions are 
actually transforming absolute errors on their inputs into relative errors in the output. 
Which means that even if the computational layers yield relatively imprecise results, activation layers 
will recover decent relative errors, as long as inputs are bounded (which is basically always the case 
thanks to bounded activation functions and normalizations).

Finally, we offered the first, to the best of our knowledge, interpretation of the impact of precision 
choice upon the top-1 accuracy of classification networks.
If we reason that the goal is to preserve the top-1 choice w.r.t. the reference model, we can establish  
the minimum required precision as a function of the deduced error bound and the distance between 
the top-1 and top-2 choice. 

This reasoning reflects perfectly the fact that as long as the model was well-trained for some 
classification problem and can clearly distinguish  between classes, then the network is extremely 
robust to low-precision evaluation.

We identify several axes of future improvements to this work.
The first improvement will be to improve the tool's performance, which so far does not scale up to 
models having tens of millions parameters. We identified the performance bottleneck to be the memory 
management in MPFI. 
The solution would be to replace the the underlying IA implementation by a faster one. 
In addition to that, the manually-coded error analysis is of course error-prone, from the formal 
verification point of view. Another limitation of the proposed tools is that it analyzes one 
implementation produced by the frugally-deep library. In order to support other implementations, e.g. 
using Kahan summation instead of a straightforward one, a corresponding code generation phase 
needs to be added. 
The second improvement concerns mixed-precision implementations, as proposed by 
NVIDIA~\cite{nvidia_mp}, which can be achieved by removing the global $\fpu$ and parameterizing the 
error analysis with the input/output precision. 
To go further, we would like to combine our results with a static analysis and mixed-precision tuning 
tool like Daisy~\cite{Daisy} to accelerate specific parts of  DNN models. 
Extension towards the training of DNNs is non-trivial and requires an analysis of gradient 
descent algorithms. 
Finally, \cite{Cohen644658} proposes to relate the classification capacity of DNNs with the 
geometry of the object manifolds issued after each layer. By combining our error analysis with the 
quantitative ``separability" measure from~\cite{Cohen644658}, we hope to back up our interpretation 
of 
the relation between precision and accuracy for DNNs with a solid theoretic-geometrical basis.


\bibliographystyle{IEEEtran}
\bibliography{accurateai}
\end{document}